\documentclass[conference]{IEEEtran}
\usepackage{verbatim}

\ifCLASSINFOpdf
\else
\fi
\usepackage{amsmath}
\usepackage{graphicx}
\usepackage[table,xcdraw]{xcolor}
\usepackage{siunitx}
\usepackage{cite}
\begin{document}
%
\title{Online Estimation of Multiple Dynamic Graphs in Pattern Sequences}

\author{\IEEEauthorblockN{Jimmy Gaudreault}
\IEEEauthorblockA{Polytechnique Montreal\\
Quebec, Canada\\
Email: jimmy.gaudreault@polymtl.ca}
\and
\IEEEauthorblockN{Arunabh Saxena}
\IEEEauthorblockA{Indian Institute of Technology\\
Bombay, India\\
Email: arunabh96@gmail.com}
\and
\IEEEauthorblockN{Hideaki Shimazaki}
\IEEEauthorblockA{Graduate School of Informatics,\\
Kyoto University\\
Honda Research Institute Japan\\
Email: h.shimazaki@kyoto-u.ac.jp}}


%


\maketitle

\begin{abstract}
Sequences of correlated binary patterns can represent many time-series data including text, movies, and biological signals. These patterns may be described by weighted combinations of a few dominant structures that underpin specific interactions among the binary elements. To extract the dominant correlation structures and their contributions to generating data in a time-dependent manner, we model the dynamics of binary patterns using the state-space model of an Ising-type network that is composed of multiple undirected graphs. We provide a sequential Bayes algorithm to estimate the dynamics of weights on the graphs while gaining the graph structures online. This model can uncover overlapping graphs underlying the data better than a traditional orthogonal decomposition method, and outperforms an original time-dependent Ising model. We assess the performance of the method by simulated data, and demonstrate that spontaneous activity of cultured hippocampal neurons is represented by dynamics of multiple graphs.
\end{abstract}



%
\IEEEpeerreviewmaketitle

\section{Introduction}
Networks of interacting elements, or graphs, describe varieties of systems, such as communication systems, biological and ecological systems, and social media systems  \cite{holme2012temporal}. Examples include people's communications via emails, infection of diseases, or communications between neurons in nervous systems. A great majority of these networks can be simplified to undirected graphs of binary units, where nodes take either the active or inactive state, and connections between two nodes do not have directionality. For example, spiking activity of neural populations has been investigated using models of binary patterns \cite{schneidman2006weak}. In addition, sequences of correlated binary patterns can represent many artificial time-series data, such as text, movies, and music \cite{graves2013generating}. 

Networks of pairwise connected elements are generally described as a Markov random field, or an undirected graphical model; with nodes and edges representing random variables and their relations respectively. This model can be written using the exponential family distribution. When the random variables are binary, their probability distribution follows the celebrated Ising model, or the Boltzmann machine-- a stochastic extension of the Hopfield memory network. However, it is unlikely that observed binary patterns are sampled from such a stationary distribution except for controlled conditions. Hence, the model was extended so it can capture time-varying structures in the data \cite{Shimazaki2009,song2009keller,kolar2010estimating,long2011statistical,kass2011assessment,Shimazaki2012,shimazaki2013single,osogami2015seven,donner2017approximate}.

Of these time-dependent Ising models, methods based on the state-space framework offer the sequential Bayes algorithm that provides estimates of the time-varying network with Bayesian credible intervals, and the expectation-maximization (EM) algorithm that optimizes parameters in the model \cite{Shimazaki2009,Shimazaki2012,shimazaki2013single,donner2017approximate}. However, while estimating a time-dependent full Ising network is an important approach in elucidating binary activity dictated by all the pairwise interactions, a few dominant patterns may capture these time-series due to the innate correlations. Capturing essential structures in the time-varying data is an important tool to perform compression, prediction, and control of these patterns. In this study, we provide an online method to decompose the underlying structure, assuming different time-scales for hierarchical structures in the binary data generation. 

Here we construct a low-dimensional representation of the Ising model that summarizes networks as a weighted combination of undirected graphs, assuming that the weights follow independent processes and that the underlying structure of the graphs changes slowly compared to the speed of weight changes. We provide a sequential Bayes algorithm to trace the dynamics of the weights, which we call network states, while updating the underlying graph structures online. The process of learning the underlying graphs is much slower than the network dynamics caused by weight changes, which allows separation of the two dynamics. The method is thus expected to extract temporal hierarchies in natural and artificial systems that generate binary patterns. This approach makes a contrast to previous offline unsupervised methods based on orthogonal or low-rank representation of the network dynamics to reduce dimensionality of time-dependent Ising models \cite{hayashi2009dynamic,hirayama2016sparse}.
 
This paper is constructed as follows. In Methods, we explain (A) the stationary Ising model of binary patterns and construct (B) a low-dimensional model composed of multiple weighted graphs. We then introduce (C) the state-space model in which the weights of the graphs change. We derive (D) the sequential Bayes algorithm for estimating the weights as well as (E) an online update algorithm of the graphs. In Results, (A) we corroborate the algorithm using simulated data, and demonstrate that the method extracts time-varying features better than a traditional orthogonal decomposition method. Next, (B) the model selection is performed to select the number of features in the data. We confirm the selected model exhibits improved goodness-of-fit over the original time-dependent Ising model. Finally, (C) we show the applicability of the method in analyzing neural data. In Discussion, we conclude the paper with possible extensions of the method.

\section{Methods}

\subsection{The Ising model}
The time-series model of binary data we propose in this study is based on the Ising/spin-glass model in statistical physics. In order to specify notations, here we describe the Ising model. This model will be extended to the multi-graph and time-dependent model in the next sections. For $N$ binary variables, the Ising model is given as
\begin{equation}
p(\mathbf{x}\vert \mathbf{j}) 
 = \exp\left[ \sum_{i=1}^N h_i x_i + \sum_{i<j} j_{ij}x_ix_j - \psi (\mathbf{j}) \right].
\label{eq:ising_original}
\end{equation}
Here the pattern $\mathbf{x}=(x_1,\ldots,x_N)^{\prime}$ is a $N$-tuple binary vector, where each entry indicates the active $x_i=1$ or inactive $x_i=0$ state of the $i^{th}$ node. Each node of the pattern can represent a pixel in image recognition or the activity of a neuron in neuroscience spiking data. $\psi(\mathbf{j})$ is a log normalization term. It ensures that the probabilities of all patterns sum up to 1. $h_i$ is the bias parameter for the $i^{th}$ node whereas $j_{ij}$ represents the interaction between the $i^{th}$ and the $j^{th}$ nodes. We express the parameters by a single column vector: 
\begin{equation}
\mathbf{j} =(h_1,\ldots,h_N,j_{1,2},\ldots,j_{N-1,N})^\prime.
\end{equation}
We call this vector the graph of the Ising model. By adjusting the graph, we can embed correlated patterns such as binary images as the ones that frequently appear in the model. The Ising model is written in the canonical form as
\begin{equation}
p(\mathbf{x}\vert \mathbf{j})=\exp\left[\mathbf{j}^{\prime}\tilde{\mathbf{F}}(\mathbf{x})-\psi(\mathbf{j}) \right],
\end{equation}
where $\tilde{\mathbf{F}}(\mathbf{x})$ is the feature vector computed from the binary variables as
\begin{equation}
\tilde{\mathbf{F}}(\mathbf{x})=(x_1,\ldots,x_N,x_1x_2,\ldots,x_{N-1}x_{N})^\prime.
\end{equation}
The log normalization is then computed as 
\begin{equation} \label{eq:freeenergy}
\psi(\mathbf{j}) = \log\sum_\mathbf{x}\exp[\mathbf{j}^{\prime}{\tilde{\mathbf{F}}(\mathbf{x})}].
\end{equation}
The computation of $\psi(\mathbf{j})$ requires considering all possible binary patterns. The first derivative of $\psi(\mathbf{j})$ with respect to $\mathbf{j}$ provides the expectation of the feature vector:
\begin{equation}
\tilde{\boldsymbol\eta} \equiv \frac{\partial\psi(\mathbf{j})}{\partial\mathbf{j}}=E_{\mathbf{X}\vert\mathbf{j}} \tilde{\mathbf{F}}(\mathbf{X}),
\label{eq:eta_tilde}
\end{equation}
where $\mathbf{X}$ is a sample of binary data, and $E_{\mathbf{X}\vert\mathbf{j}}$ represents the expectation of $\mathbf{X}$, using $p(\mathbf{x}|\mathbf{j})$. Furthermore, the second derivative yields the Fisher information matrix: 
\begin{equation}
\tilde{\mathbf{G}} \equiv \frac{\partial^2\psi(\mathbf{j})}{\partial\mathbf{j}\partial\mathbf{j}^{\prime}} = E_{\mathbf{X}\vert\mathbf{j}}\tilde{\mathbf{F}}(\mathbf{X})\tilde{\mathbf{F}}(\mathbf{X})^{\prime} - E_{\mathbf{X}\vert\mathbf{j}}\tilde{\mathbf{F}}(\mathbf{X})E_{\mathbf{X}\vert\mathbf{j}}\tilde{\mathbf{F}}(\mathbf{X})^{\prime} .
\label{eq:G_tilde}
\end{equation}
We will utilize these relations frequently in the next sections. 

\subsection{The multi-graph Ising model}
Next, we construct an Ising model composed of multiple graphs, assuming that the data is sampled from a network that is a weighted combination of these graphs. Namely, we consider the following model:
\begin{equation}
p(\mathbf{x}\vert \boldsymbol\theta,\mathbf{J}) = \exp\left[ \sum_{k=1}^D \theta^k \left[ \sum_{i=1}^N h_i^k x_i + \sum_{j>i} j_{ij}^k x_i x_j\right] - \psi(\boldsymbol\theta,\mathbf{J})\right],
\end{equation}
where $D$ is the number of undirected graphs. We call this model a multi-graph Ising model. $h_i^k$ and $j_{ij}^k$ correspond to the graph parameters of the $k^{th}$ graph. $\theta^k$ is the weight applied to the $k^{th}$ graph, and we define $\boldsymbol\theta = (\theta^1,\theta^2,\hdots,\theta^D)^{\prime}$. If $h_i^k$ and $j_{ij}^k$ are known, the weights $\theta^k$ constitute the canonical parameters of the model. Given the graph parameters and $D < N + N(N - 1)/2$, this model can represent the data with a smaller dimensionality. 
The multi-graph Ising model can be written in the canonical form as
\begin{equation}
p(\mathbf{x}\vert \boldsymbol\theta,\mathbf{J}) = \exp\left[\boldsymbol\theta^\prime \mathbf{F}(\mathbf{x}, \mathbf{J})-\psi(\boldsymbol\theta,\mathbf{J})\right].
\label{eq:low-diemnsional_ising_model}
\end{equation}
Here, the feature vector, $\mathbf{F}(\mathbf{x}, \mathbf{J})$, is given by:
\begin{equation}
\mathbf{F}(\mathbf{x}, \mathbf{J})=
\begin{bmatrix}
 \sum\limits_{i=1}^N h^1_i 		x_i+\sum\limits_{j>i} j^1_{ij} x_i x_j \\
  \sum\limits_{i=1}^N h^2_i x_i+\sum\limits_{j>i} j^2_{ij} x_i x_j \\
    \vdots \\
\sum\limits_{i=1}^N h^D_i x_i+\sum\limits_{j>i} j^D_{ij} x_i x_j
\end{bmatrix},
\end{equation}
and $\mathbf{J}$ is the matrix containing all graphs:
\begin{align}
\mathbf{J} 
&=
\begin{bmatrix}
	\mathbf{j}^1 & \mathbf{j}^2 & \hdots & \mathbf{j}^D
\end{bmatrix}.
\end{align}
Using $\mathbf{J}$, the feature vector can be further expressed as
\begin{equation}
\mathbf{F}(\mathbf{x}, \mathbf{J}) = \mathbf{J}^{\prime}\tilde{\mathbf{F}}(\mathbf{x}).
\end{equation}
Note that $\tilde{\mathbf{F}}(\mathbf{x})$ is the feature vector of the original Ising model of Eq.~\ref{eq:ising_original}. The log normalization term, $\psi(\boldsymbol\theta,\mathbf{J})$, is then calculated in the same way as in the original Ising model: $\psi(\boldsymbol\theta,\mathbf{J}) = \log\sum_\mathbf{x}\exp[\boldsymbol\theta^\prime \mathbf{F}(\mathbf{x},\mathbf{J})]$. 

Like in the original Ising model, the first and second-order derivatives of the log-normalization term respectively provide the expectation and the second-order cumulant of the new feature vector $\mathbf{F}(\mathbf{x},\mathbf{J})$. These can be expressed using the expectation parameters and the Fisher information of the original Ising model (Eqs.~\ref{eq:eta_tilde},\ref{eq:G_tilde}). The expectation of the feature vector is given as 
\begin{align}
\boldsymbol\eta
&=E_{\mathbf{X}\vert\boldsymbol\theta,\mathbf{J}}\mathbf{F}(\mathbf{x},\mathbf{J})
=\mathbf{J}^\prime E_{\mathbf{X}\vert\boldsymbol\theta,\mathbf{J}} \tilde{\mathbf{F}}(\mathbf{x})
=\mathbf{J}^\prime \tilde{\boldsymbol\eta} \nonumber \\
&=
\begin{bmatrix}
	\mathbf{j}_{1}^{\prime}\tilde{\boldsymbol\eta} &
    \mathbf{j}_{2}^{\prime}\tilde{\boldsymbol\eta} &
    \hdots &
    \mathbf{j}_{D}^{\prime}\tilde{\boldsymbol\eta} 
\end{bmatrix}^{\prime},
\label{eq:expectation_eta}
\end{align}
where $E_{\mathbf{X}\vert\boldsymbol\theta,\mathbf{J}}$ represents the expectation of $\mathbf{X}$, using $p(\mathbf{x}\vert\boldsymbol\theta,\mathbf{J})$. Similarly, the second-order cumulant of the feature vector, i.e., the Fisher information matrix $\mathbf{G}$, is given as
\begin{align}
\mathbf{G}
&=E_{\mathbf{X}\vert\boldsymbol\theta,\mathbf{J}}\left[\mathbf{F}(\mathbf{x},\mathbf{J})\mathbf{F}(\mathbf{x},\mathbf{J})^{\prime} \right] -  E_{\mathbf{X}\vert\boldsymbol\theta,\mathbf{J}}\mathbf{F}(\mathbf{x},\mathbf{J}) 
E_{\mathbf{X}\vert\boldsymbol\theta,\mathbf{J}}\mathbf{F}(\mathbf{x},\mathbf{J})^{\prime} \nonumber \\
&= E_{\mathbf{X}\vert\boldsymbol\theta,\mathbf{J}}\left[
\mathbf{J}^\prime \tilde{\mathbf{F}}(\mathbf{x})
\tilde{\mathbf{F}}(\mathbf{x})^\prime \mathbf{J} \right] 
- E_{\mathbf{X}\vert\boldsymbol\theta,\mathbf{J}}\mathbf{J}^\prime \tilde{\mathbf{F}}(\mathbf{x})
E_{\mathbf{X}\vert\boldsymbol\theta,\mathbf{J}}\tilde{\mathbf{F}}(\mathbf{x})^\prime \mathbf{J} \nonumber \\
&=\mathbf{J}^\prime 
\left\{ 
E_{\mathbf{X}\vert\boldsymbol\theta,\mathbf{J}}[\tilde{\mathbf{F}}(\mathbf{x})
\tilde{\mathbf{F}}(\mathbf{x})^\prime ] 
- E_{\mathbf{X}\vert\boldsymbol\theta,\mathbf{J}}\tilde{\mathbf{F}}(\mathbf{x})
E_{\mathbf{X}\vert\boldsymbol\theta,\mathbf{J}}\tilde{\mathbf{F}}(\mathbf{x})^\prime
\right\} 
\mathbf{J} \nonumber \\
&= \mathbf{J}^\prime \tilde{\mathbf{G}} \mathbf{J}. 
\label{eq:FisherInfoMatrix}
\end{align}

\subsection{The state-space multi-graph Ising model}

\begin{figure}
\centering
\includegraphics[width=0.4\textwidth,height=30mm]{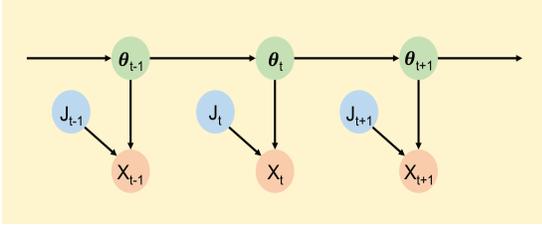}
\caption{A schematic of the state-space multi-graph Ising model. The binary pattern $\mathbf{X}_t$ is generated by multiple graphs $\mathbf{J}_t$ with their weights given by $\boldsymbol{\theta}_t$ (network state). These parameters are time-dependent.
}
\label{fig:state model}
\end{figure}

The model developed so far can only produce stationary data. In order to analyze the dynamics of binary patterns, it is essential to consider variations of model parameters in time. Figure \ref{fig:state model} shows the state-space modeling framework used for this purpose. We assume the following state equation to link the states of two consecutive time bins:
\begin{align}\label{eq:statemodel}
\boldsymbol\theta_t = \boldsymbol\theta_{t-1} + \boldsymbol{\xi}_t(\lambda).
\end{align}
At each time step, we add a zero mean Gaussian noise, $\boldsymbol\xi_t$, that has a covariance given by $\lambda^{-1}\mathbf{I}$, where $\mathbf{I}$ is a $D \times D$ identity matrix. $\lambda$ is time-independent and will be fixed. This state equation assumes that the dynamics of the network states are independent Gaussian processes.

To use this state equation, it is necessary to assume the state of the model at the first time bin. We assume its probabilistic distribution to be Gaussian, with mean $\boldsymbol\mu$ and covariance $\boldsymbol\Sigma$: $\boldsymbol\theta_1 \sim \mathcal{N}(\boldsymbol\mu,\boldsymbol\Sigma)$. We will fix these hyper-parameters ($\boldsymbol\mu$ and $\boldsymbol\Sigma$). However, unlike the algorithm developed in \cite{Shimazaki2009,Shimazaki2012,shimazaki2013single,donner2017approximate}, here we assume that one of the hyper-parameters, namely $\mathbf{J}$, is also time-dependent, and denote it as $\mathbf{J}_t$ for $t=1,\ldots,T$.

At each time step, we assume that the data is sampled from the current multi-graph Ising model with network state $\boldsymbol\theta_t$ and graphs $\mathbf{J}_t$:
\begin{align}\label{eq:observationmodel}
&p(\mathbf{x}\vert \boldsymbol\theta_t,\mathbf{J}_t) \nonumber \\
& =  \exp\left[ \sum_{k=1}^D \theta_{t}^{k} \left[ \sum_{i=1}^N h_i^{k,t} x_i + \sum_{i<j} j_{ij}^{k,t} x_i x_j\right] - \psi(\boldsymbol\theta_t,\mathbf{J}_t)\right] \nonumber \\ 
& =\exp\left[\boldsymbol\theta_t^\prime \mathbf{F}(\mathbf{x},\mathbf{J}_t)-\psi(\boldsymbol\theta_t,\mathbf{J}_t)\right].
\end{align}
This observation model is the main subject of this study. Eqs.~\ref{eq:statemodel} and \ref{eq:observationmodel} constitute the state-space model of the dynamic pattern sequences. With this augmentation, related parameters of the model ($\boldsymbol\theta_t$, $\mathbf{J}_t$, $\boldsymbol\eta_t$ and $\mathbf{G}_t$) now become time-dependent. The log normalization $\psi(\boldsymbol\theta_t,\mathbf{J}_t)$ is also time-dependent as it is a function of $\boldsymbol\theta_t$ and $\mathbf{J}_t$. We assume that the graph parameter $\mathbf{J}_t$ follows much slower dynamics than the changes of weights $\boldsymbol\theta_t$, which makes their respective effect on the binary patterns distinguishable. In the next sections, we aim to estimate $\boldsymbol\theta_t$ and $\mathbf{J}_t$ from time-varying data. The algorithm that will be developed in these sections is summarized in Table~\ref{pseudo code}. 

\begin{table}
\newcounter{cellnum}
\newcommand{\itemtab}{\stepcounter{cellnum}\thecellnum.}
\centering
\caption{Algorithm for the online estimation of the graphs}
\label{pseudo code}
\begin{tabular}{|l|}
\hline
\rowcolor[HTML]{EFEFEF} 
\itemtab\phantom{=} Set $t=1$ and initialize $\mathbf{J}_1$.                                     \\
\itemtab\phantom{=} Obtain the filter density of $\boldsymbol\theta_t$, using Eq.~\ref{eq:NR}.
\\ 
\rowcolor[HTML]{EFEFEF} 
\itemtab\phantom{=} Compute the gradient of the Q-function  
\\ 
\rowcolor[HTML]{EFEFEF} 
\phantom{===}evaluated at $\mathbf{J}_t$, using Eq.~\ref{update_J}. 
\\
\itemtab\phantom{=} Obtain $\mathbf{J}_{t+1}$, using Eq.~\ref{eq:J_update_qfunction}.                             \\ 
\rowcolor[HTML]{EFEFEF}
\itemtab\phantom{=} Advance in time: $t\rightarrow t+1$.                 \\
\itemtab\phantom{=} Repeat from step 2.
\\ \hline
\end{tabular}
\end{table}

\subsection{Sequential Bayes estimation}

In this subsection, we provide the sequential Bayes algorithm to estimate the time-dependent weights $\boldsymbol\theta_t$ online. Here, we denote the graphs up to time $t$ as $\mathbf{J}_{1:t}=\left\{\mathbf{J}_1,\mathbf{J}_2,\hdots,\mathbf{J}_t\right\}$. We use the same notation for the data up to time $t$ ($\mathbf{X}_{1:t}$).

The online estimation of the parameters is done by the recurrent construction of the filter density. The filter density using Bayes' theorem is given as
\begin{equation}\label{eq:posterior}
p(\boldsymbol{\theta}_t|\mathbf{X}_{1:t},\mathbf{J}_{1:t})
= \frac{p(\mathbf{X}_{t} | \boldsymbol{\theta}_t,\mathbf{J}_t) p(\boldsymbol{\theta}_t | \mathbf{X}_{1:t-1},\mathbf{J}_{1:t-1}) } {p(\mathbf{X}_{t} | \mathbf{X}_{1:t-1}, \mathbf{J}_{1:t})}.
\end{equation}
To obtain the second term of the numerator, we used:
\begin{equation}
p(\boldsymbol{\theta}_t | \mathbf{X}_{1:t-1},\mathbf{J}_{1:t})=p(\boldsymbol{\theta}_t | \mathbf{X}_{1:t-1},\mathbf{J}_{1:t-1}).
\label{eq:state-space_assumption}
\end{equation}
This holds because $\mathbf{J}_{t}$ is independent of $\boldsymbol{\theta}_t$ given that $\mathbf{X}_{t}$, which connects $\boldsymbol{\theta}_t$ and $\mathbf{J}_{t}$, is marginalized.
The estimate of the state at time $t$ ($\boldsymbol\theta_{t}$) given the data up to time $t-1$ is called one-step prediction. The one-step prediction density is derived from the Chapman-Kolmogorov equation 
\cite{smith2003estimating}:
\begin{align}\label{eq:chapmankolmogorov}
&p(\boldsymbol{\theta}_t|\mathbf{X}_{1:t-1},\mathbf{J}_{1:t-1}) \nonumber\\
&= \int p(\boldsymbol{\theta}_t| \boldsymbol{\theta}_{t-1}) p(\boldsymbol{\theta}_{t-1}|\mathbf{X}_{1:t-1},\mathbf{J}_{1:t-1}) d\boldsymbol{\theta}_{t-1}.
\end{align}
Because  the transition of the state is a Gaussian process, given that the filter density at $t-1$ is a Gaussian density with mean $\boldsymbol\theta_{t-1\vert t-1}$ and covariance $\mathbf{W}_{t-1\vert t-1}$, the one-step prediction density is also Gaussian with mean and covariance given by 
\begin{align}
\boldsymbol\theta_{t\vert t-1} &= \boldsymbol\theta_{t-1\vert t-1} \\
\mathbf{W}_{t\vert t-1} &=\mathbf{W}_{t-1\vert t-1}+\lambda^{-1}\mathbf{I}.
\end{align}

The filter density is a combination of the multi-graph Ising model and the one-step prediction density. Here, we approximate the posterior by a Gaussian using the Laplace method. As such, the mean of the approximated Gaussian density is identified as the mode of the posterior. We denote this mode as $\boldsymbol\theta_{t\vert t}$. It maximizes the log posterior given by
\begin{align}\label{eq:log_posterior}
f(\boldsymbol\theta_t) 
&\equiv  
\log p(\boldsymbol{\theta}_t|\mathbf{X}_{1:t},\mathbf{J}_{1:t}) \nonumber\\
&= \boldsymbol{\theta}_t^\prime \mathbf{F}(\mathbf{X}_t,\mathbf{J}_t) - \psi(\boldsymbol{\theta}_t) \nonumber \\
& - \frac{1}{2} (\boldsymbol{\theta}_t - \boldsymbol{\theta}_{t|t-1})' \mathbf{W}^{-1}_{t|t-1} (\boldsymbol{\theta}_t - \boldsymbol{\theta}_{t|t-1}) + \mathrm{const.}
\end{align}
Because of the log-concavity of both the observation and the state models, it is guaranteed that $f(\boldsymbol\theta_t)$ contains a unique mode that can be found by convex optimization techniques. Here we use the Newton-Raphson method:
\begin{equation}
\boldsymbol\theta^{\rm{new}}=\boldsymbol\theta^{\rm{old}}-[\boldsymbol\nabla \boldsymbol\nabla f(\boldsymbol\theta^{\rm{old}})]^{-1}\boldsymbol\nabla f(\boldsymbol\theta^{\rm{old}}).
\label{eq:NR}
\end{equation}
In this equation, the first and second derivatives are given by
\begin{align}
&\boldsymbol\nabla f(\boldsymbol\theta) =\mathbf{F}(\mathbf{X}_t,\mathbf{J}_t)-\boldsymbol\eta(\boldsymbol\theta)-\mathbf{W}_{t\vert t-1}^{-1}(\boldsymbol\theta-\boldsymbol\theta_{t\vert t-1}),　\\
&\boldsymbol \nabla \boldsymbol \nabla f(\boldsymbol\theta) =-\mathbf{G}(\boldsymbol\theta)-\mathbf{W}_{t\vert t-1}^{-1},
\end{align}
where $\boldsymbol\eta$ is the expectation of the feature by $p(\mathbf{x} \vert\boldsymbol\theta,\mathbf{J}_t)$ computed as in Eq.~\ref{eq:expectation_eta}. $\mathbf{G}$ is the Fisher information matrix calculated by Eq.~\ref{eq:FisherInfoMatrix}, where the expectation is performed by $p(\mathbf{x} \vert \boldsymbol\theta,\mathbf{J}_t)$. Given the mode found by the above algorithm, the filtered covariance at each time step is approximated using the Hessian of the log posterior evaluated at $\boldsymbol\theta_{t\vert t}$: 
\begin{equation}
\mathbf{W}_{t\vert t}=-[\left. \boldsymbol \nabla \boldsymbol \nabla f(\boldsymbol\theta)　\right\vert_{\boldsymbol\theta_{t|t}}　]^{-1}=[\mathbf{G}(\boldsymbol\theta_{t|t})+\mathbf{W}_{t\vert t-1}^{-1}]^{-1}. 
\end{equation}

This approximate Gaussian filter is then used to make a prediction density at $t+1$, which completes the recursion. 

\subsection{Identification of the underlying graphs}
We combine the sequential Bayes estimation of the time-dependent weight parameters with the online estimation of the graphs $\mathbf{J}_t$. Our approach is based on the maximum likelihood estimation of $\mathbf{J}_t$ using the stochastic gradient method. We optimize $\mathbf{J}_{t}$ under the principle of maximizing the marginal likelihood at time step $t$ given the past observations up to $t-1$: 
\begin{equation}
l(\mathbf{J}_{t}) \equiv \log p(\mathbf{X}_{t} \vert\mathbf{X}_{1:t-1},\mathbf{J}_{1:t-1},\mathbf{J}_{t}).
\end{equation}
This optimization will be performed by the stochastic gradient method: 
\begin{equation}
\mathbf{J}_{t+1}=\mathbf{J}_t + \epsilon \frac{\partial l (\mathbf{J}_{t})}{\partial\mathbf{J}_t},
\label{eq:J_update_logmarlikelihood}
\end{equation}
starting with a nominal initial estimate for $\mathbf{J}_1$. With this approach, we slightly update $\mathbf{J}_t$ in the direction toward the optimal parameters at time $t$ to obtain the graphs at time $t+1$, $\mathbf{J}_{t+1}$. This approach that adds the contribution of the data toward the optimal graph parameters at every time step allows us to estimate them in an online fashion. 

To employ the stochastic gradient method, we use the fact that the derivative of the marginal log-likelihood function at time step $t$ can be evaluated by the derivative of an alternative lower bound as shown below. For this goal, we introduce the lower bound of the marginal log-likelihood obtained by the following Jensen's inequality, similarly to the EM algorithm \cite{smith2003estimating}. Here, in order to search an optimal graph at time $t$, we introduce a variable $\mathbf{J}^{\ast}_t$. Then,
\begin{align}
l(\mathbf{J}^{\ast}_t)
&= \log \int p(\mathbf{X}_{t},\boldsymbol{\theta}_{t}\vert\mathbf{X}_{1:t-1} \mathbf{J}_{1:t-1},\mathbf{J}^{\ast}_t) d\boldsymbol{\theta}_{t} \nonumber\\
&=
\log \left\langle
\frac{p(\mathbf{X}_{t},\boldsymbol{\theta}_{t}\vert\mathbf{X}_{1:t-1} \mathbf{J}_{1:t-1},\mathbf{J}^{\ast}_t)}
{p(\boldsymbol\theta_{t}|\mathbf{X}_{1:t},\mathbf{J}_{1:t})}
\right\rangle_{\boldsymbol\theta_{t}|\mathbf{X}_{1:t},\mathbf{J}_{1:t}} \nonumber\\
& \geq 
\left\langle
\log \frac{p(\mathbf{X}_{t},\boldsymbol{\theta}_{t}\vert\mathbf{X}_{1:t-1} \mathbf{J}_{1:t-1},\mathbf{J}^{\ast}_t)}
{p(\boldsymbol\theta_{t}|\mathbf{X}_{1:t},\mathbf{J}_{1:t})}
\right\rangle_{\boldsymbol\theta_{t}|\mathbf{X}_{1:t},\mathbf{J}_{1:t}} \nonumber\\
&= 
\left\langle
\log p(\mathbf{X}_{t},\boldsymbol{\theta}_{t}\vert\mathbf{X}_{1:t-1} \mathbf{J}_{1:t-1},\mathbf{J}^{\ast}_t)
\right\rangle_{\boldsymbol\theta_{t}|\mathbf{X}_{1:t},\mathbf{J}_{1:t} \nonumber}
\\
&\phantom{=====}-\left\langle 
\log p(\boldsymbol\theta_{t}\vert\mathbf{X}_{1:t},\mathbf{J}_{1:t})
\right\rangle_{\boldsymbol\theta_{t}\vert\mathbf{X}_{1:t},\mathbf{J}_{1:t}}, 
\end{align}
where $\left\langle \cdot\right\rangle_{\boldsymbol\theta_t\vert\mathbf{X}_{1:t},\mathbf{J}_{1:t}}$ is the expectation by the filter density at time $t$. The second term of the last equality is the entropy of the filter density, which is not a function of $\mathbf{J}^{\ast}_t$. The first term is the expected complete data log-likelihood, a.k.a. Q-function: 
\begin{align}
q(\mathbf{J}^{\ast}_t \vert \mathbf{J}_t)&\equiv\left\langle
\log p(\mathbf{X}_{t},\boldsymbol\theta_{t}\vert\mathbf{X}_{1:t-1} \mathbf{J}_{1:t-1},\mathbf{J}^{\ast}_t)\right\rangle_{\boldsymbol\theta_{t}\vert\mathbf{X}_{1:t},\mathbf{J}_{1:t}} \nonumber \\
&=\left\langle\log p(\mathbf{X}_t\vert\boldsymbol\theta_t,\mathbf{J}^{\ast}_t) \right\rangle_{\boldsymbol\theta_{t}\vert\mathbf{X}_{1:t},\mathbf{J}_{1:t}} \nonumber \\
&\phantom{=}+\left\langle\log p(\boldsymbol\theta_t\vert\mathbf{X}_{1:t-1},\mathbf{J}_{1:t-1})\right\rangle_{\boldsymbol\theta_t\vert\mathbf{X}_{1:t-1},\mathbf{J}_{1:t}}.
\end{align}
In the second term of the last equality, we dropped $\mathbf{J}^{\ast}_{t}$, following the same argument made for Eq.~\ref{eq:state-space_assumption}. Hence, in the lower bound of the marginal log-likelihood, only the first term of the Q-function is a function of $\mathbf{J}^{\ast}_{t}$ :
\begin{equation}
q(\mathbf{J}^{\ast}_t \vert \mathbf{J}_t) = 
\langle
\boldsymbol{\theta}_t^\prime \mathbf{F}(\mathbf{X}_t, \mathbf{J}^{\ast}_t) - \psi(\boldsymbol{\theta}_t,\mathbf{J}^{\ast}_t) 
\rangle_{\boldsymbol\theta_{t}\vert\mathbf{X}_{1:t},\mathbf{J}_{1:t}} + c,
\end{equation}
where $c$ is a function independent of $\mathbf{J}^{\ast}_t$.

We note that the derivative of the marginal log-likelihood in Eq.~\ref{eq:J_update_logmarlikelihood} can be replaced with the derivative of the Q-function since we expect
\begin{equation}
\frac{\partial\log p(\mathbf{X}_{t}\vert\mathbf{X}_{1:t-1},\mathbf{J}_{1:t-1},\mathbf{J}_t)}{\partial\mathbf{J}_t}
\approx
\frac{\partial q(\mathbf{J}^{\ast}_t \vert \mathbf{J}_t)}{\partial\mathbf{J}^{\ast}_t}\Bigg\vert_{\mathbf{J}^{\ast}_t=\mathbf{J}_t}
\label{eq:equality}
\end{equation}
if the approximate posterior is close to the exact posterior. The derivation of Eq.~\ref{eq:equality} for the exact posterior is given at the end of this section. 
This equation means that, at $\mathbf{J}_t$, the optimal direction to update the graphs for maximizing the Q-function is the same as the direction that optimally maximizes the marginal log-likelihood. According to this, the online update of the underlying graphs at each time step with a gradient ascent scheme is given as:
\begin{equation}
\mathbf{J}_{t+1}=\mathbf{J}_t + \epsilon \frac{\partial q(\mathbf{J}^{\ast}_t \vert \mathbf{J}_t)}{\partial\mathbf{J}^{\ast}_t} \Bigg\vert_{\mathbf{J}^{\ast}_t=\mathbf{J}_t}.
\label{eq:J_update_qfunction}
\end{equation}
Here we obtain the derivative of the Q-function, using 
\begin{equation}
\frac{\partial\boldsymbol\theta_t^{\prime}\mathbf{F}(\mathbf{X}_t, \mathbf{J}^{\ast}_t)}{\partial\mathbf{J}^{\ast}_t}= \frac{\partial\boldsymbol\theta_t^{
\prime}\mathbf{J}^{\ast\prime}_t\tilde{\mathbf{F}}(\mathbf{X}_t)}{\partial\mathbf{J}^{\ast}_t} 
=\tilde{\mathbf{F}}(\mathbf{X}_t)\boldsymbol\theta_t^{\prime},
\end{equation} 
\begin{align}
\frac{\partial\psi(\boldsymbol\theta_t,\mathbf{J}^{\ast}_t)}{\partial\mathbf{J}^{\ast}_t} &= \frac{\partial}{\partial\mathbf{J}^{\ast}_t}\log\sum_{\mathbf{x}}\exp\left[\boldsymbol\theta_t^{\prime}\mathbf{F}(\mathbf{x}, \mathbf{J}^{\ast}_t)\right] \nonumber \\
&=\frac{\partial}{\partial\mathbf{J}^{\ast}_t}\log\sum_{\mathbf{x}}\exp\left[\boldsymbol\theta_t^{\prime}\mathbf{J}^{\ast\prime}_t\tilde{\mathbf{F}}(\mathbf{x})\right] \nonumber \\
&=\frac{\sum_{\mathbf{x}}\exp\left[\boldsymbol\theta_t^{\prime}\mathbf{F}(\mathbf{x}, \mathbf{J}^{\ast}_t)\right]\tilde{\mathbf{F}}(\mathbf{x})\boldsymbol\theta_t^{\prime}}{\sum_{\mathbf{x}}\exp\left[\boldsymbol\theta_t^{\prime}\mathbf{F}(\mathbf{x}, \mathbf{J}^{\ast}_t)\right]} \nonumber \\
&=\sum_{\mathbf{x}}\exp\left[\boldsymbol\theta_t^{\prime}\mathbf{F}(\mathbf{x}, \mathbf{J}^{\ast}_t)-\psi(\boldsymbol\theta_t,\mathbf{J}^{\ast}_t)\right]\tilde{\mathbf{F}}(\mathbf{x})\boldsymbol\theta_t^{\prime} \nonumber \\
&=E_{\mathbf{X}\vert\boldsymbol\theta_t,\mathbf{J}^{\ast}_t}\tilde{\mathbf{F}}(\mathbf{X})\boldsymbol\theta_t^{\prime}
=\tilde{\boldsymbol\eta}_t\boldsymbol\theta^{\prime}_t, 
\end{align}
where $E_{\mathbf{X}\vert\boldsymbol\theta_t,\mathbf{J}^{\ast}_t}$ represents the expectation by $p(\mathbf{x}\vert\boldsymbol\theta_t,\mathbf{J}^{\ast}_t)$. Using the above results, the derivative of the Q-function with respect to $\mathbf{J}^{\ast}_t$ is expressed as
\begin{align}\label{update_J}
\frac{\partial\ q(\mathbf{J}^{\ast}_t \vert \mathbf{J}_t)}{\partial\mathbf{J}^{\ast}_t}
=\langle(\tilde{\mathbf{F}}(\mathbf{X}_t)-\tilde{\boldsymbol\eta}_t)\boldsymbol\theta_t^{\prime}\rangle_{\boldsymbol\theta_{t}\vert\mathbf{X}_{1:t},\mathbf{J}_{1:t}}.
\end{align}
Updating $\mathbf{J}_t$ requires evaluating Eq.~\ref{update_J} at $\mathbf{J}^{\ast}_t=\mathbf{J}_t$, which requires the evaluation of the expectation with respect to the filter density at time $t$. This can be computed by sampling from the approximate Gaussian posterior distribution, $\boldsymbol\theta_{t}\sim\mathcal{N}(\boldsymbol\theta_{t\vert t},\mathbf{W}_{t\vert t})$. Finally, we sketch the equality for Eq.\ref{eq:equality} for an exact posterior density:
\begin{align}
&\frac{\partial q(\mathbf{J}^{\ast}_t \vert \mathbf{J}_t) }{\partial \mathbf{J}^{\ast}_t}\Bigg\vert_{\mathbf{J}^{\ast}_t=\mathbf{J}_t} \nonumber \\
&=\frac{\partial \left\langle\log p(\mathbf{X}_{t},\boldsymbol\theta_{t}\vert\mathbf{X}_{1:t-1},\mathbf{J}_{1:t-1},\mathbf{J}^{\ast}_t)\right\rangle_{\boldsymbol\theta_{t}\vert\mathbf{X}_{1:t},\mathbf{J}_{1:t}}}{\partial\mathbf{J}^{\ast}_t}\Bigg\vert_{\mathbf{J}^{\ast}_t=\mathbf{J}_t} \nonumber \\
&=\left\langle\left[\frac{\partial\log p(\boldsymbol\theta_{t}\vert\mathbf{X}_{1:t-1},\mathbf{X}_t,\mathbf{J}_{1:t-1},\mathbf{J}_{t})}{\partial\mathbf{J}_t}\right.\right. \nonumber \\
&\phantom{======}\left.\left.+ \frac{\partial\log p(\mathbf{X}_{t}\vert\mathbf{X}_{1:t-1},\mathbf{J}_{1:t-1},\mathbf{J}_t)}{\partial\mathbf{J}_t}\right]\right\rangle_{\boldsymbol\theta_{t}\vert\mathbf{X}_{1:t},\mathbf{J}_{1:t}} \nonumber \\
&=\int p(\boldsymbol\theta_{t}\vert\mathbf{X}_{1:t},\mathbf{J}_{1:t})\left[\frac{1}{p(\boldsymbol\theta_{t}\vert\mathbf{X}_{1:t},\mathbf{J}_{1:t})} \frac{\partial p(\boldsymbol\theta_{t}\vert\mathbf{X}_{1:t},\mathbf{J}_{1:t})}{\partial\mathbf{J}_t} \right. \nonumber \\ 
&\phantom{======}\left.
+\frac{\partial\log p(\mathbf{X}_{t}\vert\mathbf{X}_{1:t-1}, \mathbf{J}_{1:t})}{\partial\mathbf{J}_t}\right]d\boldsymbol\theta_{t} \nonumber \\
&=\int\left[\frac{\partial p(\boldsymbol\theta_{t}\vert\mathbf{X}_{1:t},\mathbf{J}_{1:t})}{\partial\mathbf{J}_t}\right. \nonumber \\
&\phantom{======}\left.+p(\boldsymbol\theta_{t}\vert\mathbf{X}_{1:t},\mathbf{J}_{1:t})\frac{\partial\log p(\mathbf{X}_{t}\vert\mathbf{X}_{1:t-1},\mathbf{J}_{1:t})}{\partial\mathbf{J}_t}\right]d\boldsymbol\theta_{t}\nonumber \\
&=0+1\cdot\frac{\partial\log p(\mathbf{X}_{t}\vert\mathbf{X}_{1:t-1},\mathbf{J}_{1:t})}{\partial\mathbf{J}_t}. 
\label{eq:equality_long}
\end{align}
This completes the stochastic gradient search of the underlying graphs under the maximum likelihood principle.

\section{Results}

\begin{figure*}[t]
\centering
\includegraphics[width=0.8\textwidth,height=60mm]{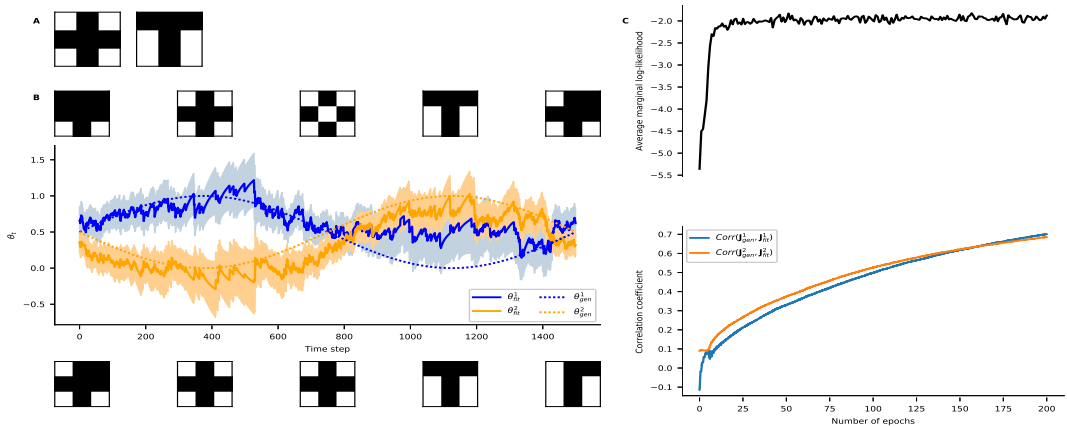}
\caption{Online estimation of graphs present in artificially generated data. The state-space multi-graph Ising model was fitted to repeatedly sampled data (200 epochs, each epoch is composed of $T_{\rm ep}=\num{1500}$ time bins). \textbf{A} Most likely binary patterns generated from two constant underlying graphs ($\mathbf{J}$) with $N=9$ nodes. \textbf{B} Up: Images sampled from the time-dependent underlying model at time steps $\left\{ \num{0}, \frac{T_{\rm ep}}{4}, \frac{T_{\rm ep}}{2}, \frac{3T_{\rm ep}}{4}, T_{\rm ep} \right\}$. Middle: Time-dependent weight parameters ($\boldsymbol\theta_t$) of the generative model (dashed line) for a single epoch. Solid lines show the maximum a posteriori estimates of the weights at the last epoch of the data with credible intervals. Bottom: Images sampled from the fitted model at the aforementioned time steps at the last epoch of the data. \text{C} Top: Average of the marginal log-likelihood of the data given the fitted model at every time step over each epoch of the data. Bottom: Correlation coefficient between the graphs of the underlying $\mathbf{J}$ and the fitted $\mathbf{J}_t$ matrices at each epoch of the data. During fitting, the variance of the columns of the estimated $\mathbf{J}_t$ matrix was scaled to unity at every time bin. $\lambda$ was fixed to 1000 and the learning rate $\epsilon$ was $10^{-3}$.}
\label{fig:gen_data}
\end{figure*}

\subsection{Application to artificially generated data}
\textbf{Generation of artificial data} To demonstrate the applicability of the method, we fitted the model to synthesized binary data representing dynamically changing images (Fig.~\ref{fig:gen_data}), in which each node ($x_i$) corresponds to a pixel that can either be black ($x_i=1$) or white ($x_i=0$). At each step, a sample was generated by Eq.~\ref{eq:observationmodel}, a mixture of multiple graphs $\mathbf{J}$ with time-dependent weights $\boldsymbol\theta_t$. Here we explain how we constructed $\mathbf{J}$ and $\boldsymbol\theta_t$ to generate data.


We generated the data from a mixture of two fixed graphs, each containing 9 nodes. We used a constant $\mathbf{J}$ to synthesize the data. If the graphs are used separately, the model most frequently generates an image similar to a `+' or a `T' respectively (Fig.~\ref{fig:gen_data}A). Note that these images have overlapped components. For the time-dependent weights, we chose sinusoidal waves of period $T_{\rm ep}$=1500 time steps. The middle panel of Fig.~\ref{fig:gen_data}B shows the dynamics of the weights of the underlying model (dashed lines) over one epoch of the data. We set their baseline and amplitude to 0.5 with a phase shift of $\pi$ so that each graph has in turn a dominant contribution while the other has a small contribution. We repeatedly generated the time-series according to the above procedure, and obtained 200 epochs of the binary data ($T$=\num{300000} time steps). 


On the top panel of Fig.~\ref{fig:gen_data}B, we show images sampled from the underlying model at different time steps in a single epoch. In particular, because of the chosen dynamics of the weights, we expect to obtain with a high probability an image corresponding to a '+' and a 'T' at $t=0.25 T_{\rm ep}$ and $t=0.75 T_{\rm ep}$ respectively. The other images are sampled from a mixture of the two graphs and do not necessarily resemble either one of the main images. 

\textbf{Estimating the state-space multi-graph Ising model} On the middle panel of Fig.~\ref{fig:gen_data}B, we also show the dynamics of the weights of the fitted model (solid lines) at the last epoch of the data. These are obtained by the sequential Bayes algorithm. We show their credible interval computed from the covariance matrix of the posterior density of the weights as
$
\boldsymbol\theta_{t\vert t}^k \pm 2\sqrt{\left[\mathbf{W}_{t\vert t}\right]_{k,k}}
$,
where $\left[\mathbf{W}_{t\vert t}\right]_{k,k}$ corresponds to the $k^{th}$ element on the diagonal of the covariance matrix. The dynamics of the fitted model seem to adequately concord with those of the underlying model, as the underlying weights are generally found in the credible intervals of the estimates. On the bottom panel of Fig.~\ref{fig:gen_data}B, we show images sampled from the fitted model for the last epoch of the data. The sampled images from the fitted model became closer to those generated artificially as the online estimation of $\mathbf{J}$ progressed. 

At a given time $t$, the marginal log-likelihood of the data can be approximated by (see \cite{Shimazaki2012}):
\begin{align}
l(\mathbf{J}_{t}) &\equiv \log p(\mathbf{X}_{t} \vert\mathbf{X}_{1:t-1},\mathbf{J}_{1:t}) \nonumber \\
&\approx \left[\boldsymbol\theta^{\prime}_{t\vert t}\mathbf{F}(\mathbf{X}_t,\mathbf{J}_t)-\psi(\boldsymbol\theta_{t\vert t},\mathbf{J}_t)\right]\nonumber \\
&-\frac{1}{2}(\boldsymbol\theta_{t\vert t}-\boldsymbol\theta_{t\vert t-1})^{\prime}\mathbf{W}^{-1}_{t\vert t-1}(\boldsymbol\theta_{t\vert t}-\boldsymbol\theta_{t\vert t-1})\nonumber \\
&+\frac{1}{2}(\log\det\mathbf{W}_{t\vert t}-\log\det\mathbf{W}_{t\vert t-1})
\end{align}
The top panel of Fig.~\ref{fig:gen_data}C shows the average of the marginal log-likelihoods obtained for each epoch. Namely, for epoch $r$, we computed:
\begin{equation}
l_{\rm avg}(r) = \frac{1}{T_{\rm ep}}\sum_{\tau=(r-1) T_{\rm ep}+1}^{r T_{\rm ep}} l(\mathbf{J}_{\tau})
\end{equation}
for $r=1,...,200$.
The marginal log-likelihood of our fitted model increased with more epochs of the data.

For this simple example, we expected that the matrix $\mathbf{J}$ that generated the data would be reconstructed by the online algorithm. In order to verify the goodness of the estimated $\mathbf{J}$, we computed the correlation coefficient between the columns of the fitted and underlying $\mathbf{J}$ matrices:  
\begin{equation}
{\rm corr. coef.} (\mathbf{j}^k_{\rm gen},\mathbf{j}^k_{\rm fit})=\frac{{\mathbf{j}^k_{\rm gen}\cdot\mathbf{j}^k_{\rm fit}} }{ {\sqrt{||\mathbf{j}^k_{\rm gen}|| \cdot ||\mathbf{j}^k_{\rm fit}||}}},
\label{eq:corr_coef}
\end{equation}
where $\mathbf{j}^k$ corresponds to the $k^{th}$ column of the $\mathbf{J}$ matrix and the dot denotes the inner product of two vectors. 
Both graphs had a correlation coefficient close to 0 at the beginning (the first guess of $\mathbf{J}$ was randomized as a zero-mean, unit-variance Gaussian) and it progressed toward 1 (Fig.~\ref{fig:gen_data}C Bottom panel). 

\begin{figure*}
\centering
\includegraphics[width=0.8\textwidth,height=60mm]{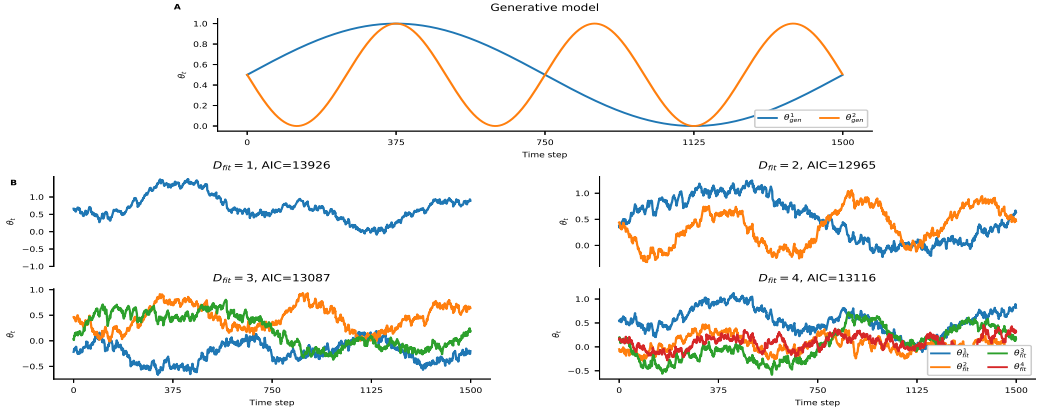}
\caption{Determination of the number of graphs present in data generated from $D$ = 2 random zero-mean, unit-variance Gaussian graphs for $N$=9 nodes. The data was generated 200 times. \textbf{A} Dynamics of the weights applied to the 2 columns of the underlying $\mathbf{J}$ matrix. \textbf{B} Dynamics of the fitted weights at the last epoch of the data for models with number of graphs ($D_{\rm{fit}}$) equal to 1, 2, 3 and 4. The Akaike Information Criterion (AIC) of each model is indicated in the plot titles. For every fitted model, the variance of the columns of the $\mathbf{J}_t$ matrix was scaled to unity at every time bin. $\lambda$ was fixed to 1000 and the learning rate $\epsilon$ was $10^{-3}$ for all fitted models.}
\label{fig:nb_graph}
\end{figure*}

Finally, in order to show that the proposed method captured the features better than a traditional orthogonal decomposition method, we performed a principle component analysis (PCA) on the MAP estimate of the time-dependent full Ising model that includes only a single graph as in Eq.~\ref{eq:ising_original}, but whose parameters all vary in time \cite{Shimazaki2009,Shimazaki2012}. We used the same binary data to fit the model. The data matrix considered to perform the PCA was constituted of the MAP estimates at every time step. Features extracted by PCA were less correlated with the generative graphs: the absolute values of the correlation coefficients between the columns of the generative $\mathbf{J}$ and PC1 and PC2 were below 0.6, which is lower than the result obtained by our method ($\sim$ 0.7). 

\subsection{Model selection}

In practical application of the method, the exact amount of graphs from which the data is sampled ($D$) is unknown. Hence, it is necessary to specify how many graphs we need to include in the model. To obtain the most predictive model that avoids overfitting to the data, we compare the models by their respective Akaike Information Criterion (AIC) given by
\begin{equation}
{\rm AIC} = -2l(\mathbf{J}_{T-T_{\rm ep}:T}) + 2m
\end{equation}
as a proxy for selection criteria of the online method. Here, $l(\mathbf{J}_{T-T_{\rm ep}:T})=\sum_{t=T-T_{\rm ep}}^T l(\mathbf{J}_t)$ and $m$ is the number of free parameters, which in our case is the number of elements in the $\mathbf{J}$ matrix. Here we compute the marginal log-likelihood of the models at the last epoch of the data. The model with the smallest AIC should be selected. 



In Fig.~\ref{fig:nb_graph}, we show that the AIC properly identified the number of graphs in artificially generated data. We generated 200 epochs of data by constructing a model containing two 9-node, zero-mean, unit-variance random Gaussian graphs. Fig.~\ref{fig:nb_graph}A shows the dynamics of the weights used to generate the data over one epoch. We then fitted the model to this data using up to 4 graphs. The same values were used for all other hyper-parameters ($\boldsymbol\mu$, $\boldsymbol\Sigma$, $\lambda$). Fig.~\ref{fig:nb_graph}B shows the fitted weight dynamics. The AIC of each model is shown in the titles of the plots. The model containing 2 graphs has the lowest AIC, which means that using the AIC as a discriminating criterion can avoid over-fitted models.

Last, we compared the goodness-of-fit of our model to that of a time-dependent full Ising model \cite{Shimazaki2009,Shimazaki2012}. For the full model, the mean of the prior of $\mathbf{j}$ was the only free parameter, so $m$ is equal to the number of elements in $\mathbf{j}$. The AIC obtained (\num{27372}) is higher than the AIC of the multi-graph models. This confirms that reducing the dimensionality can lead to a better fit to data sampled from the multi-graph Ising model. 

\subsection{Application to neural data}
\begin{figure*}[t]
\centering
\includegraphics[width=0.8\textwidth]{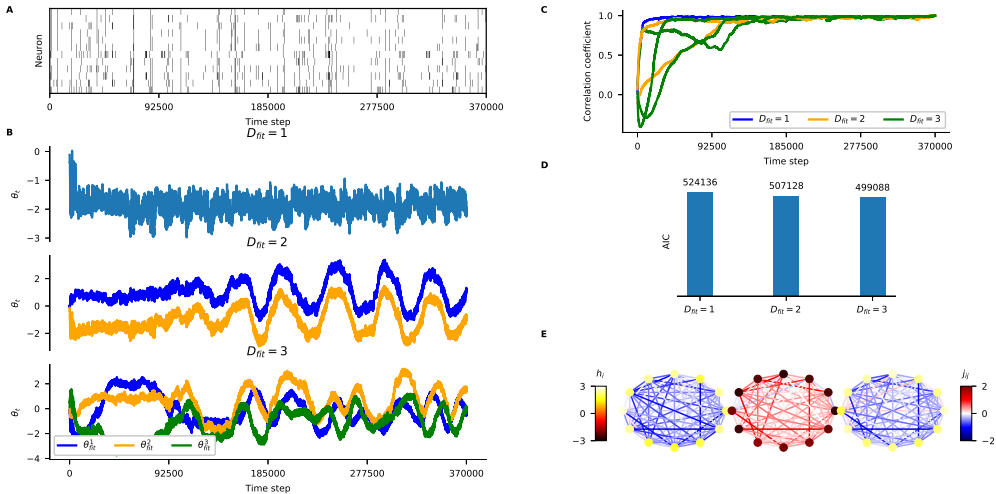} 
\caption{Application of the state-space multi-graph Ising model to neural data. \textbf{A} Binary spike patterns of neurons from rat hippocampal dissociated cultures with 10 ms bins (total $T=\num{369999}$ bins). The data comes from the 12 neurons exhibiting the highest spike rates from one neural culture (Culture 28) at day 34. \textbf{B} Dynamics of the estimated weights for models with number of graphs ($D_{\rm{fit}}$) equal to 1, 2 and 3. During the fitting procedure, the variance of the columns of $\mathbf{J}_t$ was scaled to unity at every time bin. $\lambda$ was fixed to \num{10000} and the learning rate $\epsilon$ was $10^{-2}$ for all fitted models. \textbf{C} Correlation coefficients between the columns of the estimated $\mathbf{J}$ matrix at the last time step and the columns of the estimated $\mathbf{J}$ matrix at each time step. \textbf{D} AIC of each model fitted to the last half of the data. \textbf{E} Estimated graphs at $t=T$ for the model with 3 graphs. The color of the nodes represent the values of the first-order parameters $h_i$ and the color of the edges represent the value of the second-order parameters $j_{ij}$.}
\label{fig:neural}
\end{figure*}

Next, we applied our model to neural data. We analyzed spontaneous activity of neurons recorded from rat hippocampal dissociated cultures using multi-electrode arrays \cite{10.3389/fphys.2016.00425,timme2018spontaneous}. In these experiments, the authors recorded the cultures for about 5 weeks while neurons grew connections, and reported that the neural activity approached a critical state over time \cite{10.3389/fphys.2016.00425}. The recording length is approximately 1 hour per day. 
We analyzed one culture on day 34 that contained 85 neurons. We chose the 12 neurons that showed the highest firing rates. We used 10 ms bins to construct binary data (Fig.~\ref{fig:neural}A), and fitted the multi-graph Ising models with the number of graphs ($D_{\rm{fit}}$) equal to 1, 2, and 3. Fig.~\ref{fig:neural}B shows the estimated weights $\boldsymbol \theta_{t}$ for each model. Correlation coefficients between the learned $\mathbf J$ at the last step and the graphs obtained at each step confirms that the learning of $\mathbf J$ was completed by the first half of the observation period (Fig.~\ref{fig:neural}C). The learning for models with more than 3 graphs was not completed in this period, hence these models were excluded from the analysis.
Notably, in the latter half during which $\mathbf J$ is stable, the weight of the single-graph model is stationary whereas weights of multi-graph models vary in time. Since the learned graphs within the multi-graph models ($D_{\rm{fit}}=2, 3$) are significantly correlated, the coordinated weight changes mostly capture the stationary activity of the population (not shown). Nevertheless differential dynamics of the multi-graphs may capture non-stationary activity of the population. To evaluate the predictive power of the models, we computed their AICs as its proxy, using the data in the latter half of the observation period. The model with 3 graphs was selected. This result suggests that the spontaneous activity of cultured neurons is not stationary, as reported in \cite{sasaki2007metastability}.

\section{Discussion}
We proposed an algorithm that can identify graphs underlying binary time-series data. Our model is composed of multiple time-dependent Ising graphs with weight parameters. The model is fitted by sequential Bayes estimation for the weights and stochastic gradient based on maximizing the likelihood for the graphs. The method was corroborated by artificially generated and neural time-series data. 

It should be noted that our method extracts multiple graphs whose weights follow the assumption specified by the prior density (Eq.~\ref{eq:statemodel}). In the current study, we assumed independent Markov processes for the weight dynamics. Given the observation model without the prior, the choice of $\mathbf{J}$ is undetermined because the following observation model with $D$ graphs is identical for any choice of invertible $D$ by $D$ matrix $\mathbf{Z}$:
\begin{equation}
p(\mathbf{x}\vert\boldsymbol\theta_t, \mathbf{J}_t)=\exp\left[\boldsymbol\theta_t^{\prime}(\mathbf{Z}^{-1}\mathbf{Z})^{\prime} \mathbf{J}_t^{\prime}\tilde{\mathbf{F}}(\mathbf{x})-\psi(\mathbf{Z} \boldsymbol\theta_t,\mathbf{J}_t \mathbf{Z}^{-1})\right].
\end{equation}
Because of the chosen prior distribution, the current method tends to extract independent weight processes with the same time-scale. Despite this assumption about the processes, the determination of the graphs still suffers ambiguity. 
For example, a single graph with a constant weight is captured by multiple graphs with significantly correlated weight dynamics if a linear combination of the graphs can reconstruct the original graph, as seen in the analysis of neural data. 
In future works, we need to impose different time-scales for weight processes and/or regularize $\mathbf{J}$ to disambiguate the graph estimation.

The current model can be extended in several ways toward predicting complex time-series data. First, the number of binary units that can be treated by the model needs to be increased. The exact computations of $\psi$, $\tilde{\boldsymbol\eta}$ and $\tilde{\mathbf{G}}$ using Eqs.~\ref{eq:freeenergy}, \ref{eq:eta_tilde} and \ref{eq:G_tilde} are only practically realizable with a limited number of nodes (up to approximately 20). For this goal, we can use the pseudolikelihood and TAP/Bethe approximation methods for the state-space Ising model introduced in \cite{donner2017approximate}. Second, we can use the particle filter to model a non-Gaussian posterior as long as the dimension of the state in this model is small. Finally, an important extension is to model non-linear transitions of the state and capture long-term relationships between time bins, similarly to the modern recurrent neural network models \cite{graves2013generating}.

\section{Conclusion}
In summary, the proposed model can estimate undirected graphs and their underlying dynamics in binary data in an online manner. This model is a dynamic probabilistic network that aims to predict sequences of binary patterns, assuming that they are sampled from a mixture of Ising graphs. While further development is necessary for practical use, the model potentially provides a probabilistic approach for pattern prediction that can be an alternative to classical recurrent networks.

\section*{Acknowledgment}
The authors thank Prof. Shinsuke Koyama for helpful discussions. 


\bibliographystyle{IEEEtran}


\begin{thebibliography}{10}
\providecommand{\url}[1]{#1}
\csname url@samestyle\endcsname
\providecommand{\newblock}{\relax}
\providecommand{\bibinfo}[2]{#2}
\providecommand{\BIBentrySTDinterwordspacing}{\spaceskip=0pt\relax}
\providecommand{\BIBentryALTinterwordstretchfactor}{4}
\providecommand{\BIBentryALTinterwordspacing}{\spaceskip=\fontdimen2\font plus
\BIBentryALTinterwordstretchfactor\fontdimen3\font minus
  \fontdimen4\font\relax}
\providecommand{\BIBforeignlanguage}[2]{{%
\expandafter\ifx\csname l@#1\endcsname\relax
\typeout{** WARNING: IEEEtran.bst: No hyphenation pattern has been}%
\typeout{** loaded for the language `#1'. Using the pattern for}%
\typeout{** the default language instead.}%
\else
\language=\csname l@#1\endcsname
\fi
#2}}
\providecommand{\BIBdecl}{\relax}
\BIBdecl

\bibitem{holme2012temporal}
P.~Holme and J.~Saramäki, ``Temporal networks,'' \emph{Physics Reports}, vol.
  519, no.~3, pp. 97 -- 125, 2012.

\bibitem{schneidman2006weak}
E.~Schneidman, M.~J. Berry, R.~Segev, and W.~Bialek, ``Weak pairwise
  correlations imply strongly correlated network states in a neural
  population,'' \emph{Nature}, vol. 440, no. 7087, pp. 1007--1012, 2006.

\bibitem{graves2013generating}
A.~Graves, ``Generating sequences with recurrent neural networks,''
  \emph{arXiv:1308.0850}, 2013.

\bibitem{Shimazaki2009}
H.~Shimazaki, S.-i. Amari, E.~N. Brown, and S.~Gr\"{u}n, ``State-space analysis
  on time-varying correlations in parallel spike sequences,'' \emph{ICASSP
  2009. IEEE International Conference on Acoustics, Speech and Signal
  Processing}, 2009.

\bibitem{song2009keller}
L.~Song, M.~Kolar, and E.~P. Xing, ``Keller: estimating time-varying
  interactions between genes,'' \emph{Bioinformatics}, vol.~25, no.~12, pp.
  i128--i136, 2009.

\bibitem{kolar2010estimating}
M.~Kolar, L.~Song, A.~Ahmed, and E.~P. Xing, ``Estimating time-varying
  networks,'' \emph{Ann Appl Stat}, pp. 94--123, 2010.

\bibitem{long2011statistical}
J.~D.~I. Long and J.~M. Carmena, ``A statistical description of neural ensemble
  dynamics,'' \emph{Front Comput Neurosci}, vol.~5, p.~52, 2011.

\bibitem{kass2011assessment}
R.~E. Kass, R.~C. Kelly, and W.-L. Loh, ``Assessment of synchrony in multiple
  neural spike trains using loglinear point process models,'' \emph{Ann Appl
  Stat}, vol.~5, no.~2B, p. 1262, 2011.

\bibitem{Shimazaki2012}
H.~Shimazaki, S.-i. Amari, E.~N. Brown, and S.~Gr\"{u}n, ``State-space analysis
  of time-varying higher-order spike correlation for multiple neural spike
  train data.'' \emph{PLoS Comput Biol}, vol.~8, no.~3, p. e1002385, 2012.

\bibitem{shimazaki2013single}
H.~Shimazaki, ``Single-trial estimation of stimulus and spike-history effects
  on time-varying ensemble spiking activity of multiple neurons: a simulation
  study,'' in \emph{J Phys Conf Ser}, vol. 473, no.~1, 2013, p. 012009.

\bibitem{osogami2015seven}
T.~Osogami and M.~Otsuka, ``Seven neurons memorizing sequences of alphabetical
  images via spike-timing dependent plasticity,'' \emph{Scientific reports},
  vol.~5, p. 14149, 2015.

\bibitem{donner2017approximate}
C.~Donner, K.~Obermayer, and H.~Shimazaki, ``Approximate inference for
  time-varying interactions and macroscopic dynamics of neural populations,''
  \emph{PLoS Comput Biol}, vol.~13, no.~1, p. e1005309, 2017.

\bibitem{hayashi2009dynamic}
K.~Hayashi, J.-I. Hirayama, and S.~Ishii, ``Dynamic exponential family matrix
  factorization,'' in \emph{PAKDD}, 2009, pp. 452--462.

\bibitem{hirayama2016sparse}
J.-i. Hirayama, A.~Hyv{\"a}rinen, and S.~Ishii, ``Sparse and low-rank matrix
  regularization for learning time-varying markov networks,'' \emph{Mach
  Learn}, pp. 1--32, 2016.

\bibitem{smith2003estimating}
A.~C. Smith and E.~N. Brown, ``Estimating a state-space model from point
  process observations,'' \emph{Neural Comput}, vol.~15, no.~5, pp. 965--991,
  2003.

\bibitem{10.3389/fphys.2016.00425}
N.~M. Timme, N.~J. Marshall, N.~Bennett, M.~Ripp, E.~Lautzenhiser, and J.~M.
  Beggs, ``Criticality maximizes complexity in neural tissue,'' \emph{Frontiers
  in Physiology}, vol.~7, p. 425, 2016.

\bibitem{timme2018spontaneous}
\BIBentryALTinterwordspacing
------, ``Spontaneous spiking activity of thousands of neurons in rat
  hippocampal dissociated cultures,'' \emph{CRCNS.org}, 2016. [Online].
  Available: \url{http://dx.doi.org/10.6080/K0PC308P}
\BIBentrySTDinterwordspacing

\bibitem{sasaki2007metastability}
T.~Sasaki, N.~Matsuki, and Y.~Ikegaya, ``Metastability of active ca3
  networks,'' \emph{Journal of Neuroscience}, vol.~27, no.~3, pp. 517--528,
  2007.

\end{thebibliography}
%



\end{document}